\documentclass[conference]{IEEEtran}
\usepackage{cite}

\ifCLASSINFOpdf
\usepackage[pdftex]{graphicx}
\else
\fi
\usepackage{amsmath}
\usepackage{amsfonts} 
\usepackage{amsthm}
\usepackage{amssymb}
\usepackage[boxruled]{algorithm2e}
\usepackage{url}

\usepackage{bbm}

\usepackage{kbordermatrix}

\newcommand{\pfun}{\mathop{\hbox{$\to$\kern-7pt\raise.9pt\hbox{\scalebox{1}[.55]{$|$}}\kern4pt} }}

\usepackage{array}

\begin{document}

\title{
Maneuver Identification Challenge}

\author{\IEEEauthorblockN{
Kaira Samuel, Vijay Gadepally, David Jacobs, Michael Jones, Kyle McAlpin, \\ Kyle Palko, Ben Paulk, Sid Samsi, Ho Chit Siu, Charles Yee, Jeremy Kepner
\\
\IEEEauthorblockA{MIT
}}}
\maketitle

\begin{abstract}
AI algorithms that identify maneuvers from trajectory data could play an important role in improving flight safety and pilot training.  AI challenges allow diverse teams to work together to solve hard problems and are an effective tool for developing AI solutions.  AI challenges are also a key driver of AI computational requirements.  The Maneuver Identification Challenge hosted at maneuver-id.mit.edu provides thousands of trajectories collected from pilots practicing in flight simulators, descriptions of maneuvers, and examples of  these maneuvers performed by experienced pilots.  Each trajectory consists of positions, velocities, and aircraft orientations normalized to a common coordinate system.  Construction of the data set required significant data architecture to transform flight simulator logs into AI ready data, which included using a supercomputer for deduplication and data conditioning.  There are three proposed challenges.  The first challenge is separating physically plausible (good) trajectories from unfeasible (bad) trajectories.  Human labeled good and bad trajectories are provided to aid in this task.  Subsequent challenges are to label trajectories with their intended maneuvers and to assess the quality of those maneuvers.
\end{abstract}


%
\IEEEpeerreviewmaketitle

\section{Introduction}
\let\thefootnote\relax\footnotetext{
This material is based upon work supported by the Assistant Secretary of Defense for Research and Engineering under Air Force Contract No. FA8702-15-D-0001, National Science Foundation CCF-1533644, and United States Air Force Research Laboratory Cooperative Agreement Number FA8750-19-2-1000. Any opinions, findings, conclusions or recommendations expressed in this material are those of the author(s) and do not necessarily reflect the views of the Assistant Secretary of Defense for Research and Engineering, the National Science Foundation, or the United States Air Force. The U.S. Government is authorized to reproduce and distribute reprints for Government purposes notwithstanding any copyright notation herein.
}

AI challenges developed by AI experts are a primary tool for advancing AI and engaging the AI ecosystem on important problems.   AI challenges are also a key driver of AI computational requirements.  Challenges such as YOHO~\cite{yoho}, MNIST~\cite{mnist}, HPC Challenge~\cite{hpcc}, Graph Challenge \cite{samsi2017static,kao2017streaming,kepner2019sparse}, ImageNet~\cite{imagenet} and VAST~\cite{vast1,vast2} have played important roles in driving progress in fields as diverse as machine learning, high performance computing, and visual analytics. YOHO is the Linguistic Data Consortium database for voice verification systems and has been a critical enabler of speech research. The MNIST database of handwritten letters has been a bedrock of the computer vision research community for two decades. HPC Challenge has been used by the supercomputing community to benchmark and test the largest supercomputers in the world as well as stimulate research on the new parallel programming environments. Graph Challenge has led to the development of award winning software and hardware \cite{kepner2016mathematical,davis2019algorithm,bulucc2017design}.
\begin{figure}[htb]
  	\centering
    	\includegraphics[width=\columnwidth]{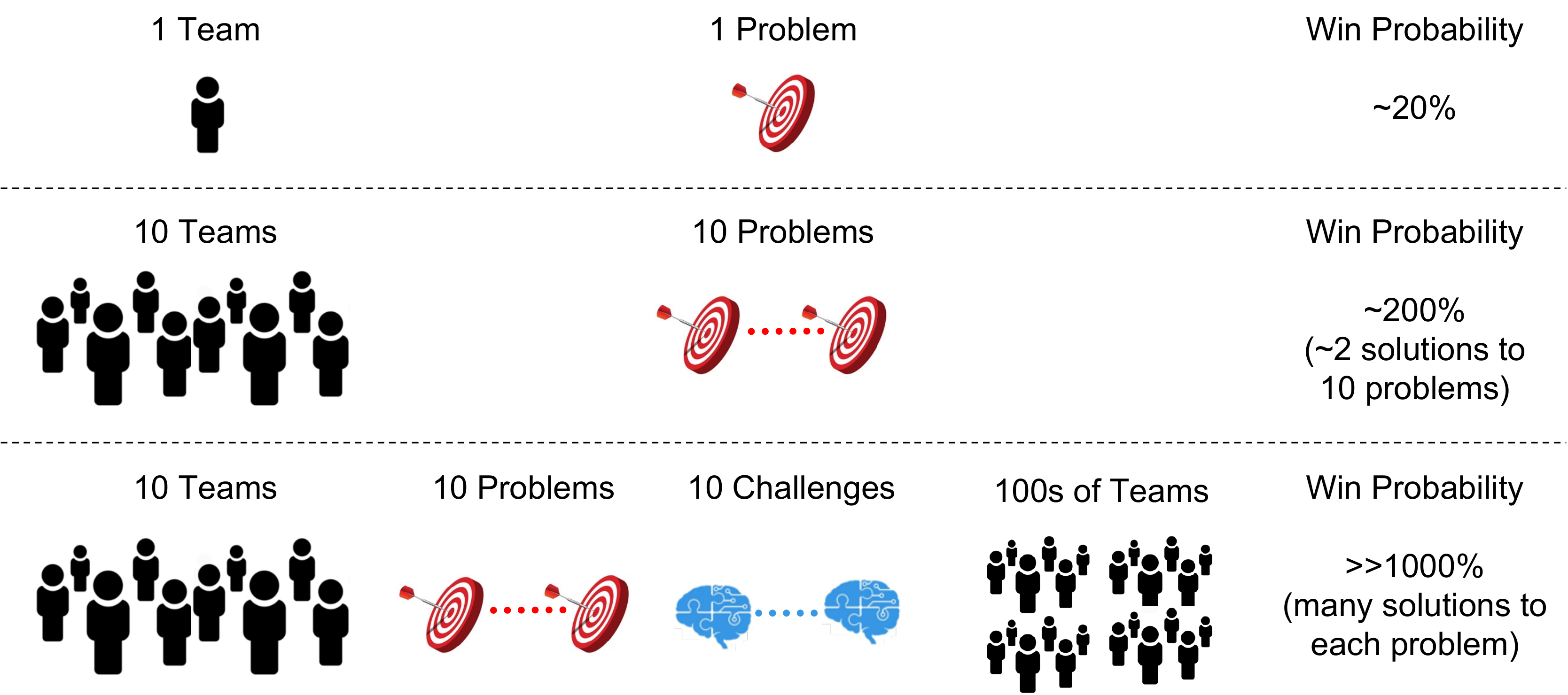}
	\caption{Challenging AI problems are best solved by enabling diverse teams to work on them.  The MIT Air Force AI Accelerator teams are creating many challenges to engage the entire AI ecosystem and maximize the number of solutions.}
      	\label{fig:ChallengeMotivation}
\end{figure}
ImageNet populated an image dataset according to the WordNet hierarchy consisting of over 100,000 meaningful concepts (called synonym sets or synsets)~\cite{imagenet}, with an average of 1000 images per synset, and has become a critical enabler of vision research. The VAST Challenge is an annual visual analytics challenge that has been held every year since 2006; each year, VAST offers a new topic and submissions are processed like conference papers.  With more AI teams working on more challenges, the gathered solutions significantly increase the probability of success (Figure~\ref{fig:ChallengeMotivation}). 

  The Maneuver Identification Challenge is part of the MIT Air Force AI Accelerator.  The AI Accelerator is designed to make fundamental advances in artificial intelligence to improve Department of the Air Force operations while also addressing broader societal needs. AI Accelerator research involves interdisciplinary teams, including Air Force personnel, who collaborate across disparate fields of AI to create new algorithms, technologies, and solutions.  A key element of the AI Accelerator is the production of AI challenges to engage the broader AI community and to institutionalize AI data readiness (aia.mit.edu/challenges).  The AI Accelerator includes challenges for optimal scheduling of training \cite{CDO2020}, magnetic navigation \cite{gnadt2020signal}, and weather prediction \cite{veillette2020sevir}.  A principal benefit of AI challenges is creating AI ready data on behalf of a community.  Transforming raw logs or observations into AI ready data is often the majority of the effort in an AI project.  A key aspect of becoming an effective AI organization is the regular production of AI ready data.

\begin{table*}
\caption{Example time, position, velocity, and orientation trajectory data for a pilot training session from a flight simulator.}
\centering
\includegraphics[width=1.5\columnwidth]{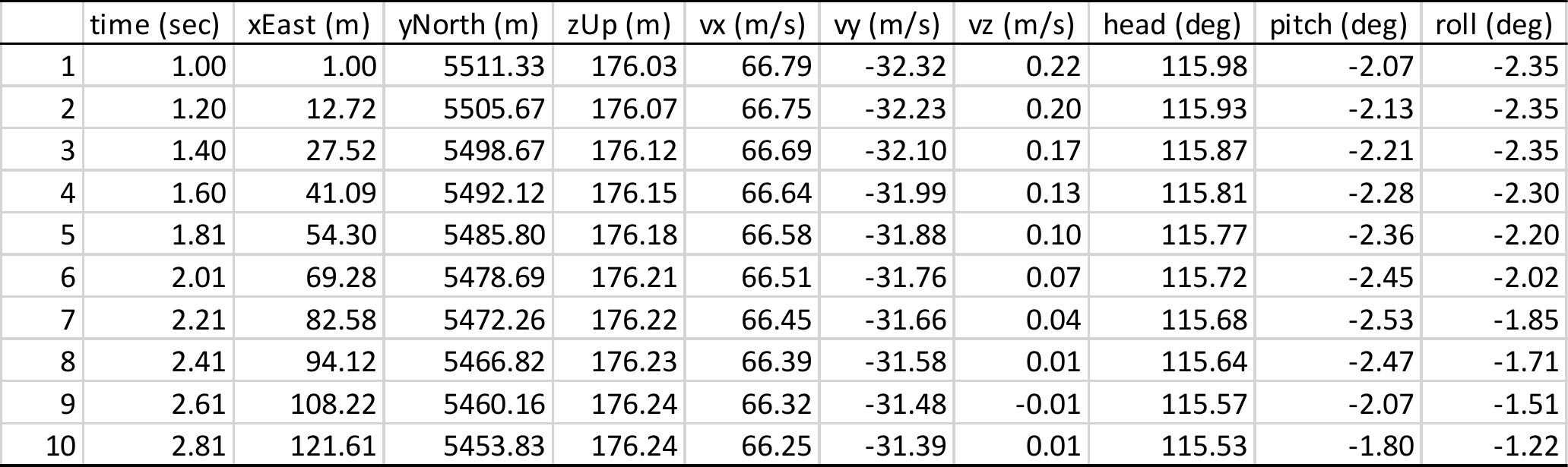}
\label{tab:TSVexample}
\end{table*}

AI algorithms that identify maneuvers from trajectory data could play an important role in improving flight safety and pilot training.  The Maneuver Identification (ID) Challenge has the potential to enhance and personalize simulator training by providing mid-flight and post-flight feedback to students.  The Air Force continues to face a pilot shortage.  Nascent technologies to individualize and automate some aspects of training present an opportunity to mitigate that shortage.  The Air Force began Pilot Training Next (PTN) to explore the use of nascent technologies such as virtual reality flight simulators and novel instructional tools in small test cohorts for personalizing training and improving student access to training resources  \cite{SAFCOPTNtechsummary}.  Specifically, the value of each flight hour in a training aircraft or virtual reality simulator can be improved by providing additional teaching tools to instructors. For example, automating flight maneuver recognition and recommending grades for each maneuver would make each debrief more efficient and effective. On the other hand, fully-automated basic maneuver recognition and grading may enable simple flight training for all levels of potential pilot candidates prior to arriving at formal pilot training. 

Maneuver identification from trajectory data is a significant technical challenge.  It is difficult to  simulate realistic flight tracks of simple maneuvers, such as beelines, racetracks, circles, s-curves, figure eights, and dives \cite{6643201}.  Determination of physical feasibility of simulated tracks presents its own challenges \cite{ure2009feasible}, as does identifying types of aircraft from tracks \cite{huang2018aircraft}.  A logistically difficult, but practical solution to obtaining realistic trajectory data is to collect it from real pilots in flight simulators.  Working with PTN, maneuver-id.mit.edu has compiled thousands of trajectories collected from pilots practicing in flight simulators, descriptions of maneuvers, and examples of these maneuvers performed by experienced pilots.  Each trajectory consists of times, positions, velocities, and aircraft orientations normalized to a common coordinate system.  Construction of the data set required significant data architecture to transform flight simulator logs into AI ready data, which included using a supercomputer for deduplication and data conditioning.  These data can be use to support a variety of challenges.  The first challenge is separating physically plausible (good) trajectories from unfeasible (bad) trajectories.  Human labeled good and bad trajectories are provided to aid in this task.  Subsequent challenges are to label trajectories with their intended maneuvers and to assess the quality of those maneuvers.

\section{Data}

The Maneuver Identification Challenge must provide a significant amount of data to train an AI. PTN has gathered thousands of distinct pilot training sessions from hundreds of hours on flight simulators, flown by students, trained pilots, and instructors.  The sessions were flown using the Lockheed Martin Prepar3d flight simulator software (prepar3d.com).  Commercial flight simulators of this type are not designed to dynamically emit data in a multi-pilot training context.  PTN has developed a novel state-of-the-art data logging system that aggregates flight simulator data from multiple simultaneous students.

\begin{figure}[htb]
  	\centering
    	\includegraphics[width=\columnwidth]{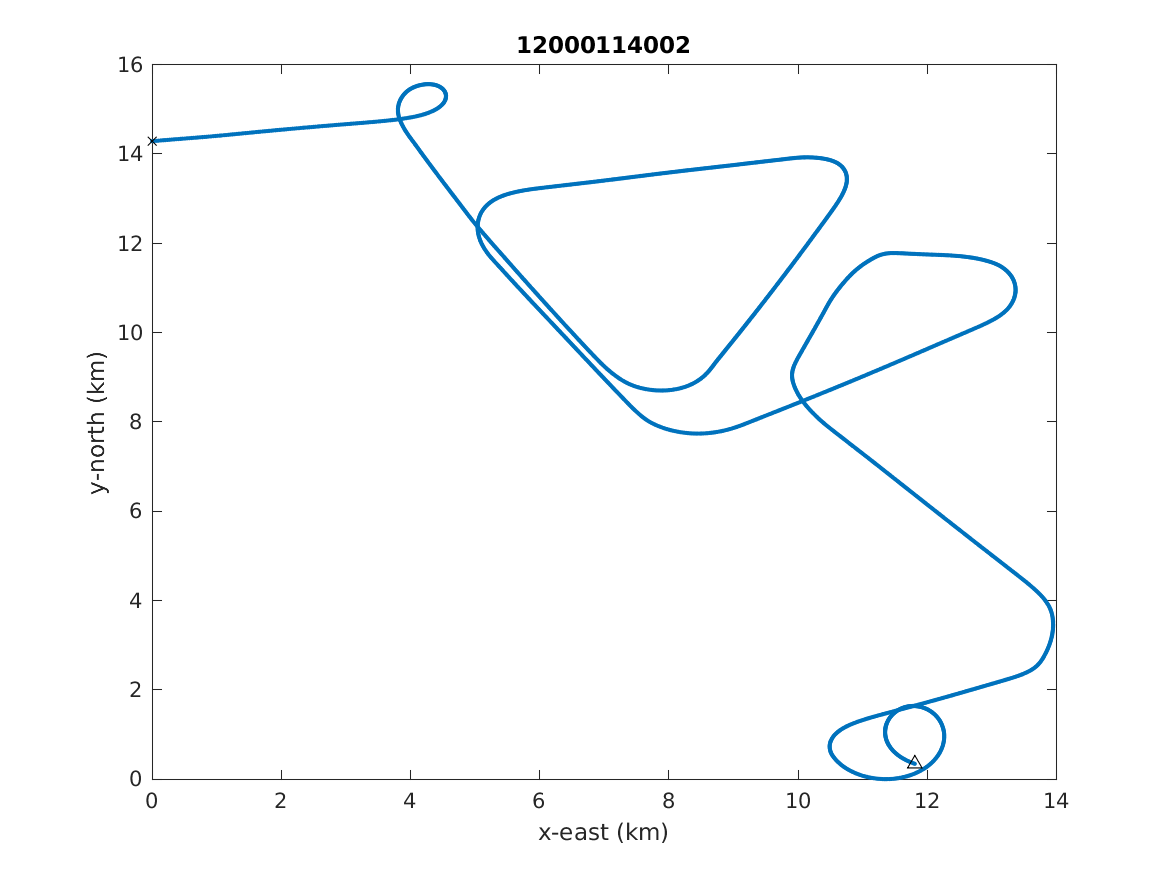}
	    \includegraphics[width=\columnwidth]{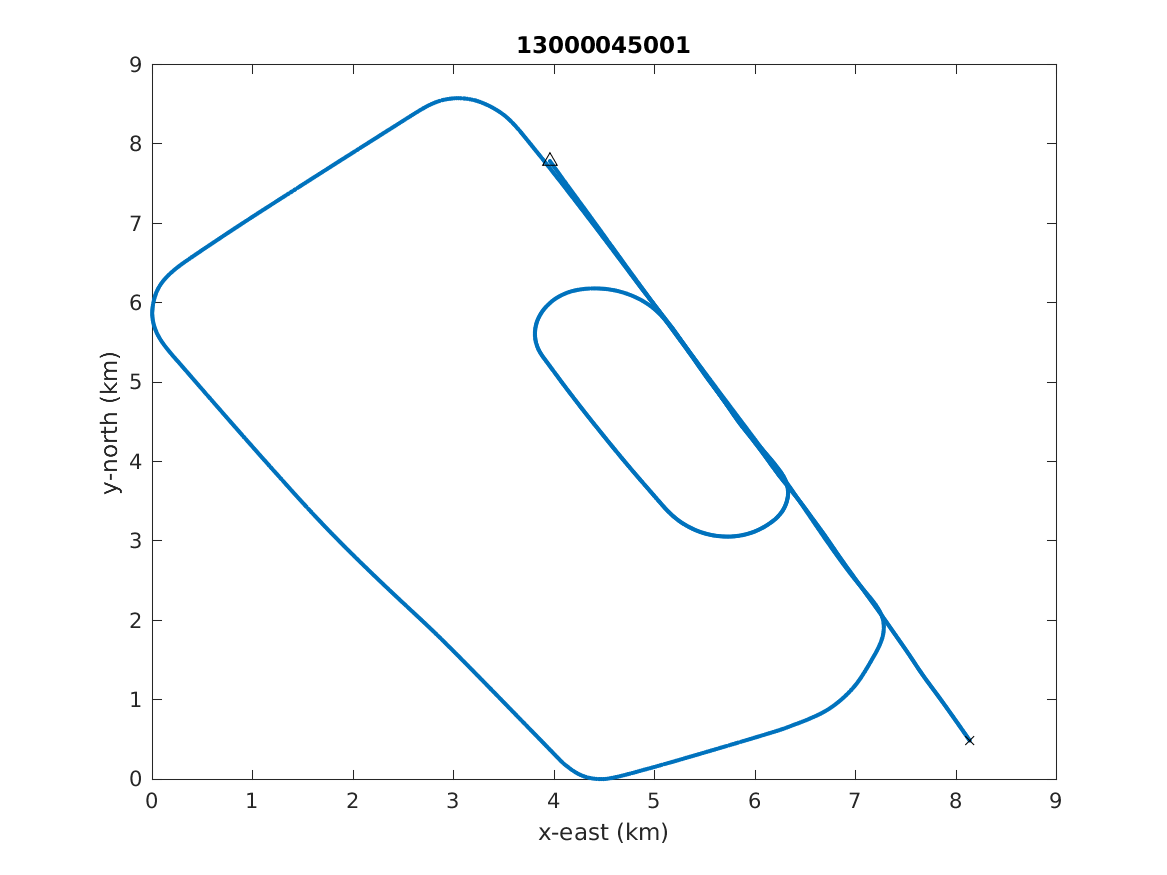}
	\caption{Examples of PNG files showing an aerial view of the flight trajectory during the session.  Trajectories begin with a $\times$ symbol and end with a $\bigtriangleup$.}
      	\label{fig:PNGexample}
\end{figure}

In general, log data is not AI ready, and requires significant transformation.  The PTN log data is stored in diverse SQL database tables that were periodically check-pointed onto cloud storage and then transferred to the MIT SuperCloud \cite{reuther2018interactive}.  The SQL checkpoints were launched using the MIT SuperCloud interactive database management system \cite{prout2015enabling}.  Each checkpoint consisted of hundreds of tables, some containing millions of records, with dozens of distinct columns.  Determination of the relevant data was accomplished by extracting the data in SQL tables using the D4M system \cite{gadepally2018d4m} and performing a subsequent dimensional data analysis \cite{gadepally2014bigdata}.

The extracted trajectory data consists of co-blended information from many pilot sessions.  For redundancy, the logging system created duplicate entries and data from idle sessions.  Deduplication involved extracting each session and then doing a full pairwise comparison of every record with every other record in the session. Using 256 64-core compute nodes (16,384 cores total) on the MIT SuperCloud all the data could be deduplicated in a few hours.  The code was implemented using Matlab/Octave with the pMatlab parallel library \cite{Kepner2009}.  A typical run could be launched in a few seconds using the MIT SuperCloud triples-mode hierarchical launching system \cite{reuther2018interactive}.  Typical launch parameters were [256 16 4], corresponding to 256 nodes, 16 Matlab/Octave processes per node, and 4 OpenMP threads per process.  On each node, the 16 processes were pinned to 4 adjacent cores to minimize interprocess contention and maximize cache locality for the OpenMP threads \cite{byun2019optimizing}.  Each Matlab/Octave process was assigned a subset of the data to deduplicate. Within each Matlab/Octave process, the underlying OpenMP parallelism was used on 4 cores to accelerate the processing.  After deduplication, sessions with less than 1000 contiguous valid entries (approximately 200 seconds) were excised, and the data was normalized and saved as tabular data files for each session.

The trajectory data has been made available through maneuver-id.mit.edu as tab separated value (TSV) files.  In addition, top-down views of the trajectories are stored as portable network graphics (PNG) files. The TSV files consist of a plain text table containing the times, positions, velocities, and orientations of the aircraft throughout the flight session. The files can be read by most data processing systems and viewed in any spreadsheet program. Table~\ref{tab:TSVexample} shows an example of a TSV file for a single session. The PNG files contain images with an aerial view of the position of the aircraft in kilometers during the session. Figure~\ref{fig:PNGexample} shows examples of a PNG file for two sessions.

The data is being shared using standard trusted data sharing best practices \cite{kepner2021zero}
\begin{itemize}
\item Data is made available in curated repositories
\item Using standard anonymization methods where needed: hashing, sampling, and/or simulation
\item Registration with a repository and demonstration of legitimate research need
\item Recipients legally agree to neither repost a corpus nor deanonymize data
\item Recipients can publish analysis and data examples necessary to review research
\item Recipients agree to cite the repository and provide publications back to the repository
\item Repositories can curate enriched products developed by researchers
\end{itemize}
In accordance with the above practices, the trajectory data has been anonymized and a data sharing agreement is required by researchers who want to obtain the data to confirm the legitimacy of the research and to ensure that they agree to respect the anonymity.  Once participants confirm the agreement they gain access to the data. Currently, there are two data sets containing thousands of files provided by PTN labeled 12000000000 and 13000000000, as well as close to 30 data files from sessions conducted at Vance Air Force Base. More may be added in the future.

\begin{figure}[htb]
  	\centering
	    \includegraphics[width=0.75\columnwidth]{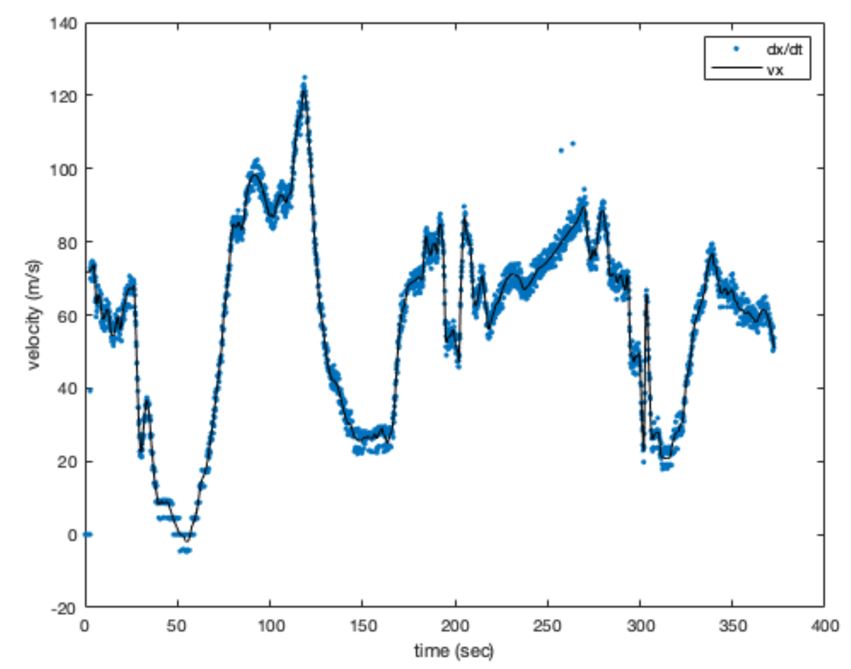}
	    \includegraphics[width=0.75\columnwidth]{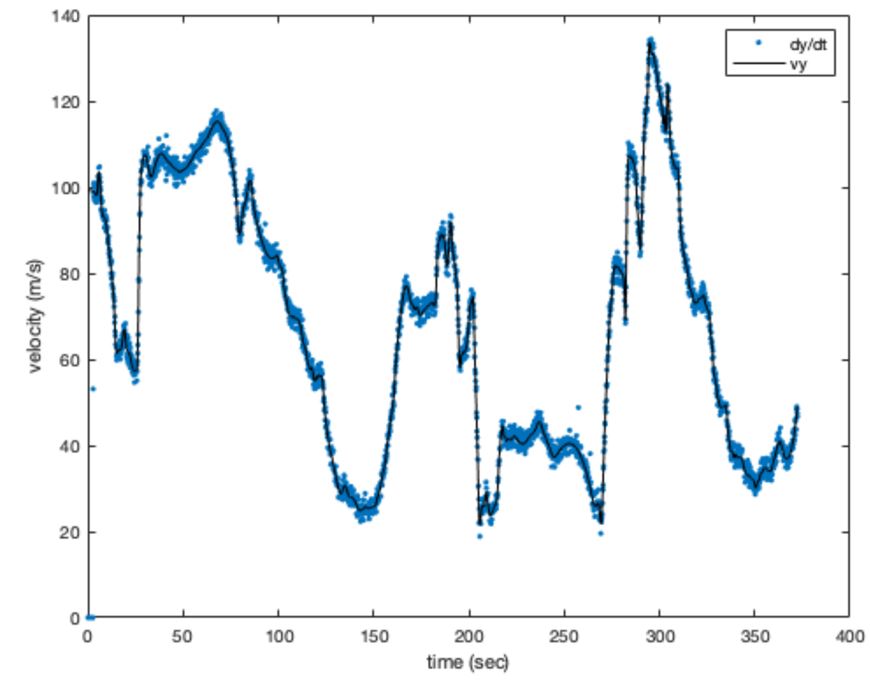}
    	\includegraphics[width=0.75\columnwidth]{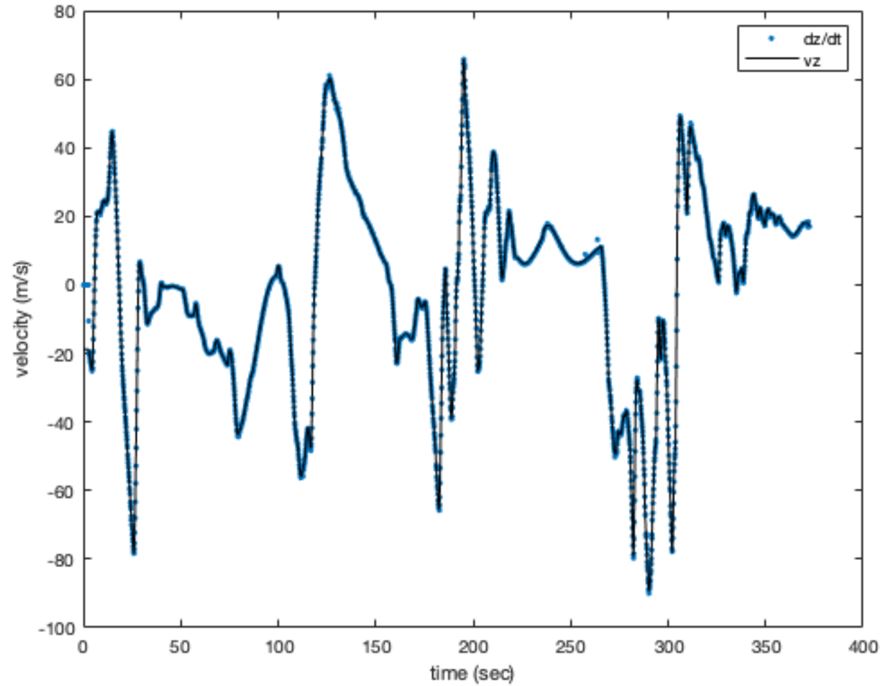}
	\caption{Plots comparing the (vx,vy,vz) velocities with the derivatives (dx/dt,dy/dt,dz/dt) computed from the (x,y,z) positions of the trajectory data.}
      	\label{fig:DataverPlot}
\end{figure}

\subsection{Normalization}

The data has been normalized to make it accessible to the broadest range of AI researchers. Each session has been set to start at one second because some data processing systems treat 0 as empty. For the same reason, the minimum spatial coordinate values for each session has been set to (1,1,1). All sessions were placed in a common (x,y,z) Cartesian coordinate system. The longitude, latitude, and altitude have been converted to units of meters. The conversion was done with MATLAB using the {\tt wgsEllipsoid} and {\tt Geodetic2enu} functions.  The following mathematical conversions were done to set the minimum coordinate values
\begin{verbatim}
     xEast  = xEast  - min(xEast)  + 1
     yNorth = yNorth - min(yNorth) + 1
     zUp    = zUp    - min(zUp)    + 1
\end{verbatim}
A caveat of the altitude normalization is that some maneuvers are altitude dependent, which is not recoverable from this normalization.  The velocities are in units of meters/second and the aircraft orientations are in degrees.  

\subsection{Verification}
  Flight simulators use a variety of coordinate systems that need to be deduced from the data. For example, some simulators may define `y' as up and others may define `z' as up.  Converting from longitude, latitude, and altitude into (x,y,z) Cartesian coordinates allows for cross validation of coordinates, velocities, and orientations.  The velocities (vx,vy,vz) were verified by comparing with the derivatives (dx/dt,dy/dt,dz/dt) computed from the (x,y,z) positions (Figure~\ref{fig:DataverPlot}).  The heading was verified by comparing with
$$
  {\rm tan}^{-1}({\rm vx/vy})
$$
Likewise, the pitch was verified by comparing with
$$
  {\rm tan}^{-1}\left(\frac{{\rm vz}}{\sqrt{{\rm vx}^2 + {\rm vy}^2}}\right)
$$

\section{Maneuvers}

\begin{figure}[htb]
  	\centering
    	\includegraphics[width=\columnwidth]{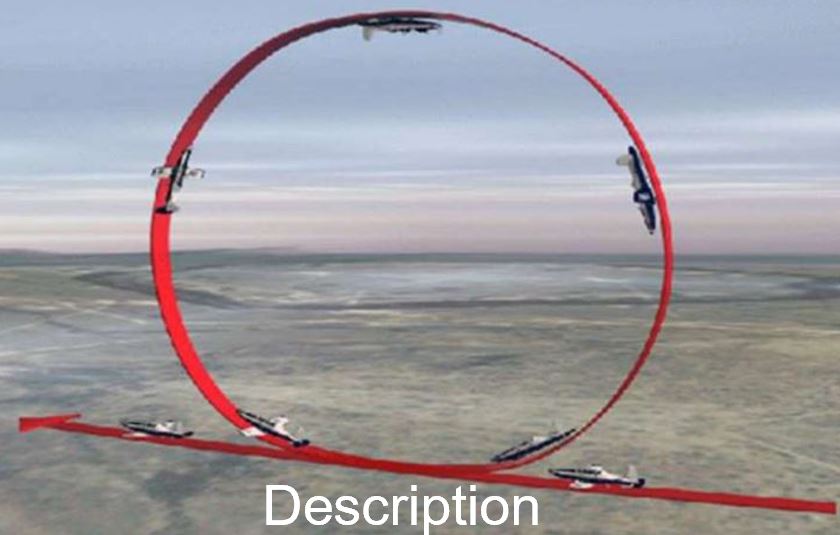}
	\caption{Sample description photo of a loop maneuver (reproduced from \cite{AETC-11-248}).}
      	\label{fig:SampleDescPhoto}
\end{figure}
\begin{figure}[htb]
  	\centering
    	\includegraphics[width=\columnwidth]{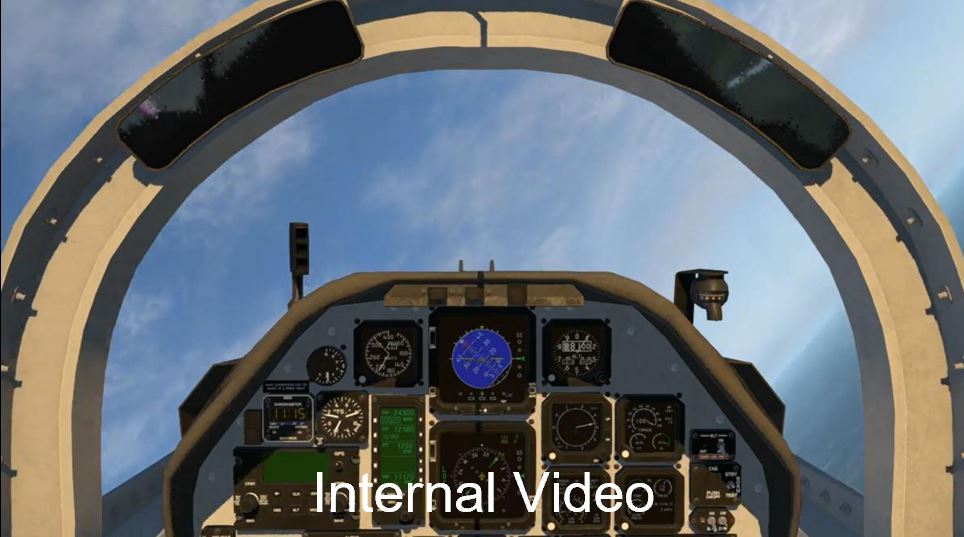}
	\caption{Screenshot from a video of an experienced pilot flying in the simulator taken from the vantage point of the virtual cockpit.}
      	\label{fig:SampleIntVid}
\end{figure}
\begin{figure}[htb]
  	\centering
    	\includegraphics[width=\columnwidth]{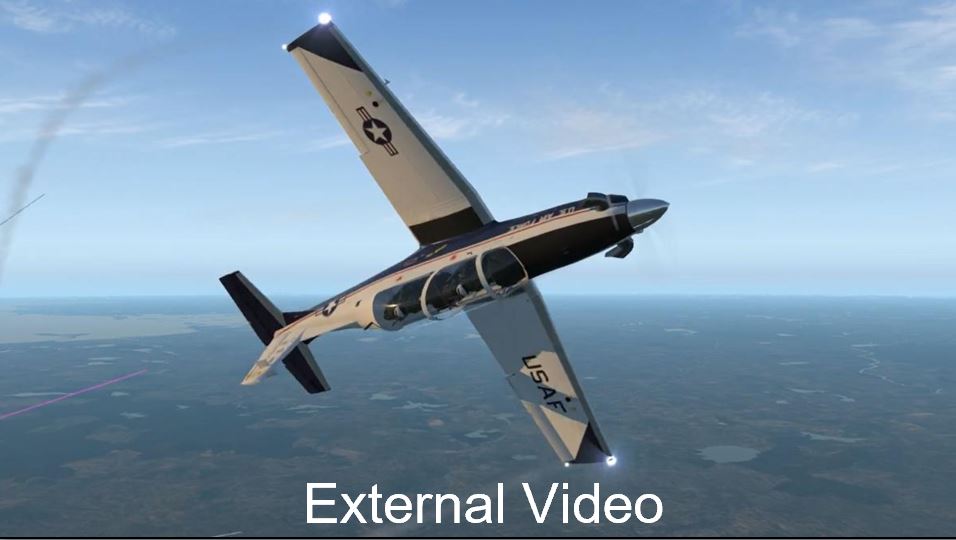}
	\caption{Screenshot from a video of an experienced pilot flying in the simulator taken from the vantage point of an outside observer.}
      	\label{fig:SampleExtVid}
\end{figure}
\begin{figure}[htb]
  	\centering
    	\includegraphics[width=\columnwidth]{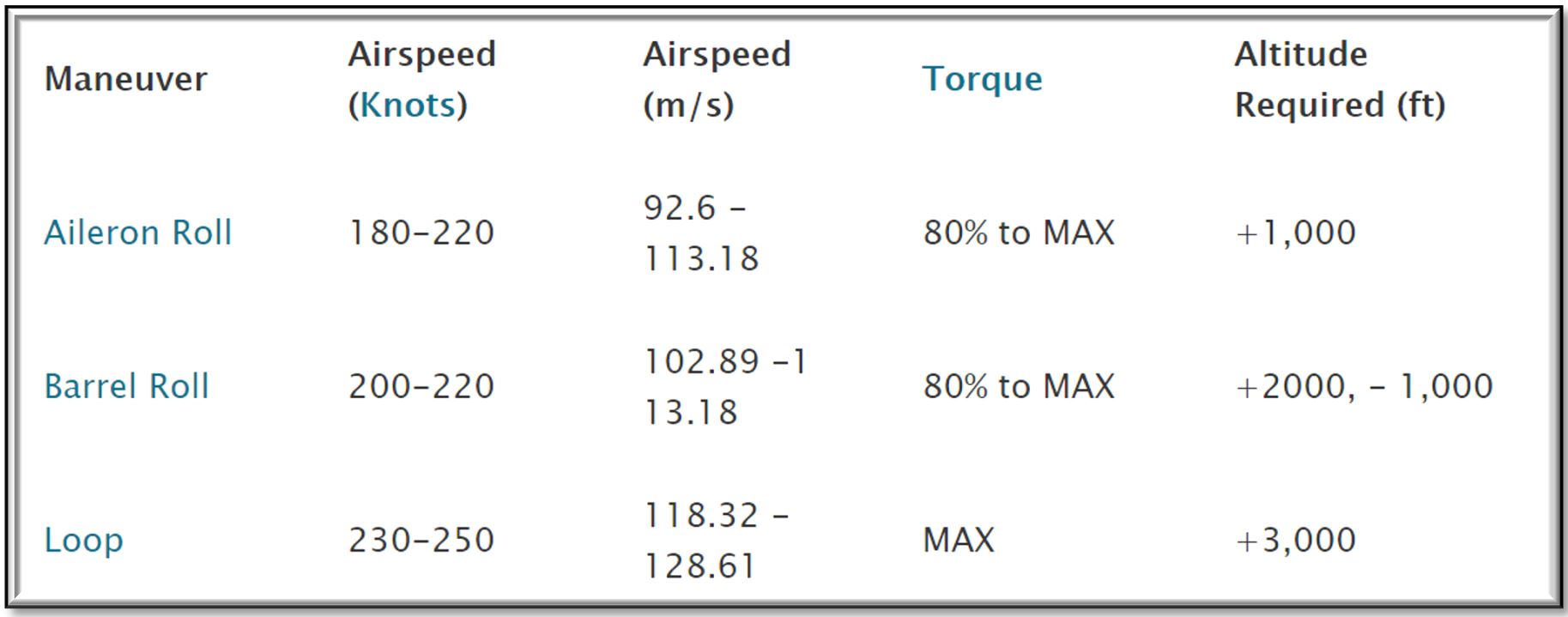}
	\caption{Sample table listing various maneuvers. The airspeed is in both knots, the standard for aviators, and meters/second, the standard for scientists. The torque settings are provided for additional context. The altitude required gives the altitude needed to execute the maneuver, where '+' indicates 'above' and '-' indicates 'below'. For example, the Barrel Roll requires 2,000 feet of clearance above and 1,000 feet below.}
      	\label{fig:ManeuverTable}
\end{figure}

Maneuver identification requires a representative set of maneuvers for AI researchers to draw upon. A  database of maneuvers has been compiled drawing from pilot training documentation \cite{AETC-11-248}. This database contains maneuver descriptions for 18 categories of maneuvers detailing both the trajectory of the maneuver and how to execute it as a pilot. Sample trajectory data, as well as photos and videos of expert pilots in flight simulations, have also been made available.

Maneuver-id.mit.edu provides details on common pilot training maneuvers. For selected maneuvers, there are trajectory descriptions (Figure~\ref{fig:SampleDescPhoto}) and how-to instructions, as well as external and internal videos of experienced pilots performing the maneuvers. Viewers can watch experienced pilots performing maneuvers from the pilot vantage point inside the simulator cockpit (Figure~\ref{fig:SampleIntVid}) as well as from a third person view outside of the aircraft (Figure~\ref{fig:SampleExtVid}). There is also trajectory data for these maneuvers in the form of TSV plain text tables and PNG files. These maneuvers are organized in a table that lists all of the maneuvers with links to descriptions and examples (Figure~\ref{fig:ManeuverTable}).

\begin{figure*}[htb]
  	\centering
    	\includegraphics[width=1.5\columnwidth]{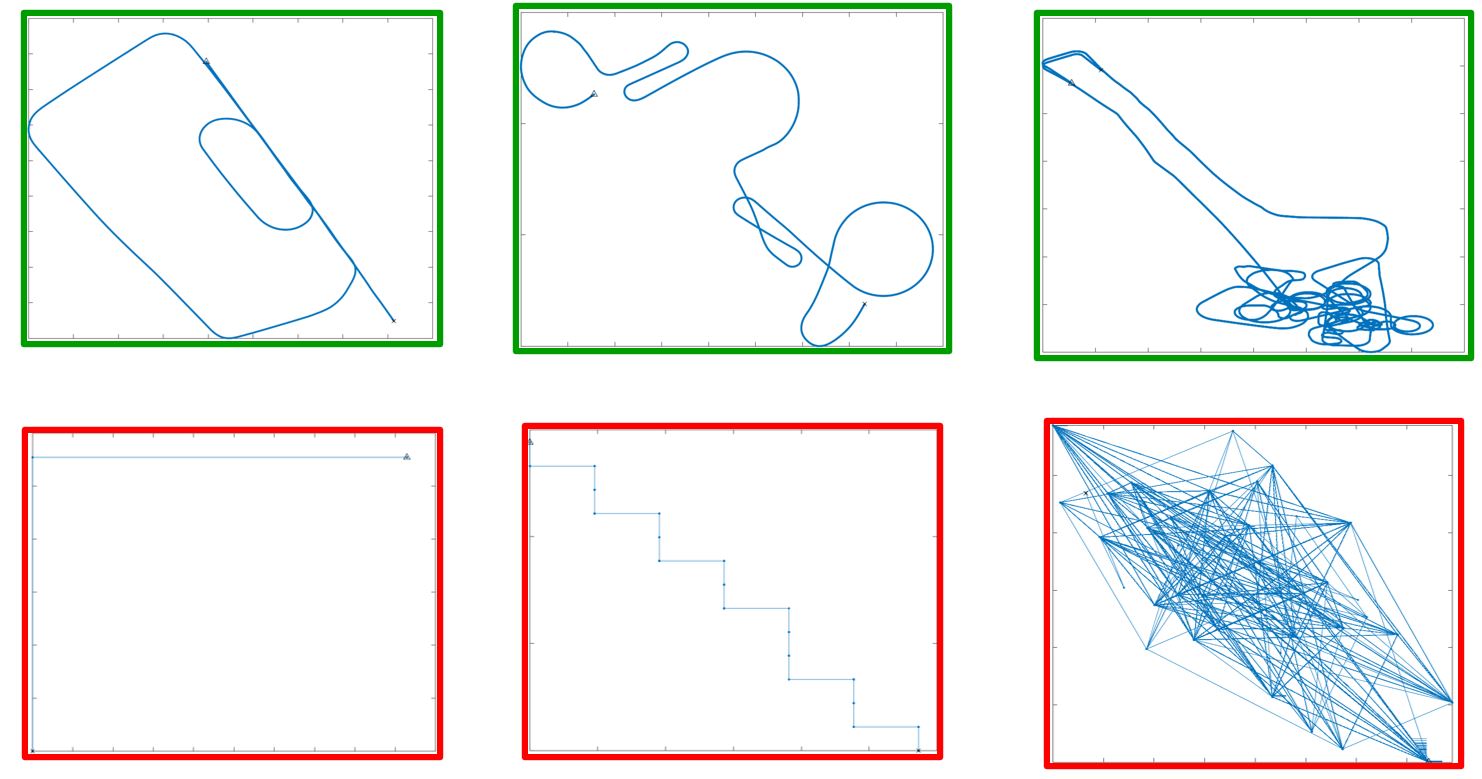}
	\caption{Examples of good data (green) and bad data (red), sorted manually. The bad data includes straight lines, jumps, and impossible maneuvers.}
      	\label{fig:Challenge1}
\end{figure*}

\section{Challenges}

PTN has begun the process of creating and testing algorithms for maneuver detection and scoring.  In their initial work, PTN researchers have focused on loops because of their discernible trajectory both visually and based on time series data parameters \cite{SAFCOPTNtechsummary}. The broad methodology consists of comparing the flight simulator data to a convolution kernel of a known maneuver example, and determining whether the two sets matched.  The convolution kernel was obtained through the virtual instructor pilot that PTN developed, and the selected example is a demonstration loop and its mirror image. Matching the flight simulator data to the kernel, each row is compared and 'time-matched', the distance between the time-matched points is found and corrected for variability, and the total distance is acquired by summing up the individual distances. This value is a measure of similarity between the flight data and the maneuver. While these methods were generally successful, more could be done to identify and account for false negatives and false positives.

  Similar work has been done outside PTN to identify maneuvers which can provide researchers looking to participate in the challenge with sample approaches. One such project was motivated by the importance of maneuver detection for load analyses on  aircraft.  Researchers analyzed the flight data and compared it to a library of maneuvers \cite{WANG2015133}. Other methods of maneuver identification, such as neural networks, have also been studied \cite{RODIN199295}. Neural networks are attractive  because of their fault tolerance, which means the occasional wrong or approximated data point will not skew the results too significantly. Neural networks have been used for classification of underwater sonar tracks and radar tracks, and for target track estimation \cite{RODIN199295,GORMAN198875,farhat1986optical,farhat1989echo}. 
  
  Automated maneuver grading is another objective of this challenge. The current method of student grading is labor-intensive and subjective as it relies on a very limited number of instructor pilots manually taking notes on each maneuver performed by a student. Differences in instructor pilot techniques, preferences, and grading inject some subjectivity into the grading \cite{SAFCOPTNtechsummary}. As the number of students increases, it is harder for instructor pilots to give each student proper attention for them to improve and learn. With automated maneuver scoring, novice student pilots could achieve a basic level of proficiency through automated feedback and grading, then rely on instructor pilots for the more advanced ``art'' of flying each maneuver. PTN has outlined a general methodology on maneuver scoring which would require having time series data and accounting for co-variance between the selected trajectory dimensions \cite{PTNv2}. The methodology consists of three main components: characterizing each dimension's behavior, characterizing each dimension's variability, and combining the first two into a single score. Scoring would consist of finding a best fit function and the corresponding acceptable variance according to expert pilot data in comparison with student runs.

Based on this prior work, maneuver-id.mit.edu has identified three challenges for the community. The first challenge deals with sorting the physically feasible (good) and physically infeasible (bad) data into separate sets based on the presence of unbroken trajectories (good) and identifiable maneuvers (good) and straight lines (bad), jumps (bad), and impossible maneuvers (bad). Examples of these good and bad data are shown in Figure~\ref{fig:Challenge1}. Currently, there is good and bad trajectory truth data that has been sorted and verified manually. The manually sorted truth data is included as part of the data set. The second challenge deals with identifying which maneuver the pilot is attempting to execute. Truth data for this challenge is not currently available and may be released later. The third challenge deals with scoring the pilot once the maneuver has been identified. This could greatly benefit the efficiency and quality of the pilot training education. Truth data for this challenge is not currently available and may be released later.

\section{Summary}

AI challenges and data sets are important tools for applying AI to pilot training and offer many potential  benefits. Effectively identifying and assessing maneuvers from aircraft data is one such challenge. Maneuver-id.mit.edu has compiled a multitude of resources that can be used to construct relevant challenges. These include flight simulator data containing positions, velocities, and orientations; classifications of good and bad simulator data; maneuver descriptions; and flight simulator data and videos of example maneuvers performed by experienced pilots. The AI community can use these resources to highlight their innovations for classifying good and bad trajectory data, identifying maneuvers, and assessing maneuvers.


\section*{Acknowledgments}
%
%

The authors wish to acknowledge the following individuals for their contributions and support: Sean Anderson, Ross Allen, Bob Bond, Jeff Gottschalk, Tucker Hamilton, Chris Hill, Mike Kanaan, Tim Kraska, Charles Leiserson, Mimi McClure, Christian Prothmann, John Radovan, Steve Rejto, Daniela Rus, Allan Vanterpool, Marc Zissman, and the MIT SuperCloud team: Bill Arcand, Bill Bergeron, David Bestor, Chansup Byun, Nathan Frey, Michael Houle, Matthew Hubbell, Hayden Jananthan, Anna Klein, Joseph McDonald, Peter Michaleas, Julie Mullen, Andrew Prout, Antonio Rosa, Albert Reuther, Matthew Weiss.



\bibliographystyle{ieeetr}
\bibliography{ManeuverIDBib}
%

\appendices
\end{document}